\documentclass[letterpaper]{article} 
\usepackage{aaai2026}  
\usepackage{times}  
\usepackage{helvet}  
\usepackage{courier}  
\usepackage[hyphens]{url}  
\usepackage{graphicx} 
\usepackage{natbib}  
\usepackage{caption} 
\usepackage{algorithm}
\usepackage{algorithmic}
\usepackage{amsmath}
\usepackage{amssymb}
\usepackage{booktabs}
\usepackage{xcolor}
\usepackage{multirow}
\usepackage{amsthm}
\usepackage{newfloat}
\usepackage{listings}
\DeclareCaptionStyle{ruled}{labelfont=normalfont,labelsep=colon,strut=off} 
\floatstyle{ruled}
\newfloat{listing}{tb}{lst}{}
\floatname{listing}{Listing}
\title{Space Alignment Matters: The Missing Piece for Inducing Neural Collapse in Long-Tailed Learning}

\author{
    Jinping Wang,
    Zhiqiang Gao\thanks{Corresponding author: zgao@wku.edu.cn.},
    Zhiwu Xie
}
\affiliations{
    College of Science, Mathematics and Technology, Wenzhou-Kean University\\
    \{1306325, zgao, zxie\} @wku.edu.cn
}

\usepackage{bibentry}

\begin{document}

\maketitle

\begin{abstract}
Recent studies on Neural Collapse (NC) reveal that, under class-balanced conditions, the class feature means and classifier weights spontaneously align into a simplex equiangular tight frame (ETF). 
In long-tailed regimes, however, severe sample imbalance tends to prevent the emergence of the NC phenomenon, resulting in poor generalization performance.
Current efforts predominantly seek to recover the ETF geometry by imposing constraints on features or classifier weights, yet overlook a critical problem: There is a pronounced misalignment between the feature and the classifier weight spaces. 
In this paper, we theoretically quantify the harm of such misalignment through an optimal error exponent analysis.
Built on this insight, we propose three explicit alignment strategies that plug-and-play into existing long-tail methods without architectural change. 
Extensive experiments on the CIFAR-10-LT, CIFAR-100-LT, and ImageNet-LT datasets consistently boost examined baselines and achieve the state-of-the-art performances.

\end{abstract}



\section{Introduction}

\begin{figure}[!t]
\centering

\parbox{1\linewidth}{
    \centering
    \includegraphics[width=1\linewidth]{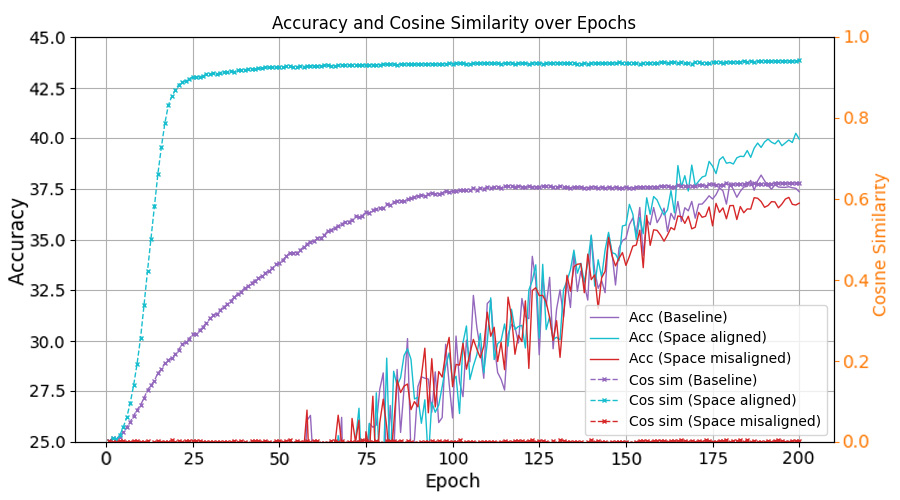}\\
    (a) 
}
\parbox{0.32\linewidth}{
    \centering
    \includegraphics[width=\linewidth]{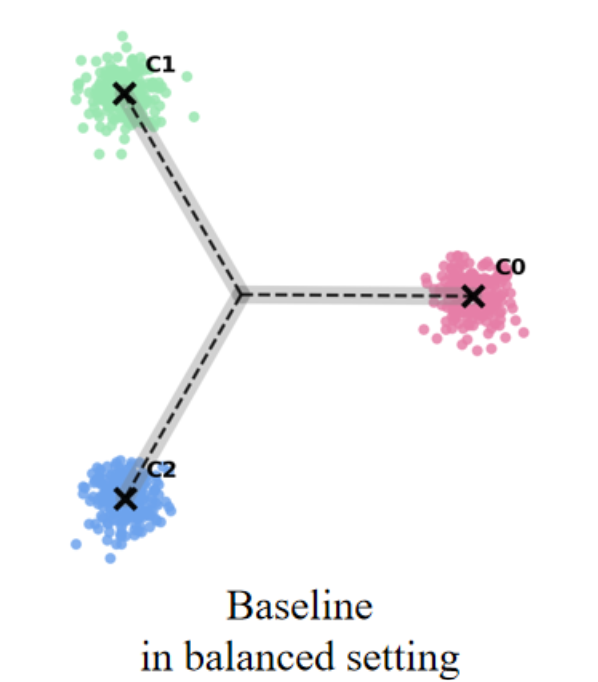}\\
    (b) 
}
\hfill
\parbox{0.32\linewidth}{
    \centering
    \includegraphics[width=\linewidth]{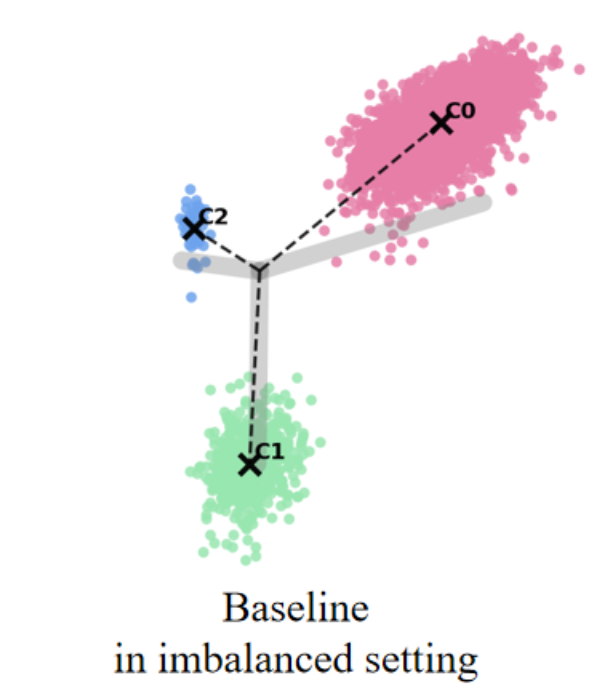}\\
    (c)  
}
\hfill
\parbox{0.32\linewidth}{
    \centering
    \includegraphics[width=\linewidth]{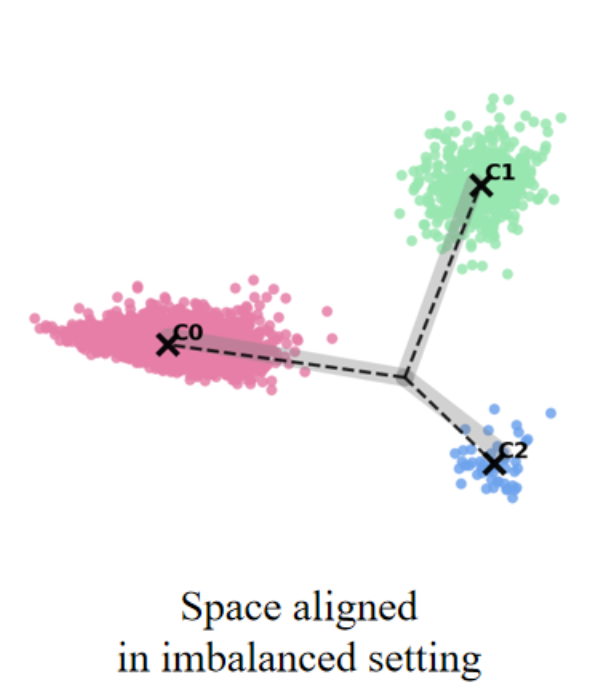}\\
    (d)  
}
\caption{Space misalignment issue.
(a) shows ResNet-32 model performances and corresponding cosine similarities between class feature means and classifier weights on the CIFAR-100 dataset. 
The baseline, aligned, and misaligned models are obtained by training with cross-entropy (CE), SpA-Reg (proposed method for alignment), and negative SpA-Reg loss, respectively.
(b), (c) and (d) are toy examples with 2-dimensional features and 3 classes, which illustrate different geometric structures of the class feature means (black crosses) and the classifier weights (gray lines). 
}
\label{fig:toy_examples}
\end{figure}

Long-tailed learning refers to scenarios where a few head classes dominate the dataset, while numerous tail classes have scarce examples. This distribution poses a significant challenge for neural networks, often resulting in suboptimal feature learning, biased predictions toward head classes, and poor generalization on minority classes \cite{Miniority_collapse}.
In contrast, when trained on balanced data, as shown in Figure~\ref{fig:toy_examples}(b), deep models exhibit the phenomenon of \textit{Neural Collapse} (NC) \cite{Neural_Collapse}, where the learned features and classifier converge to a highly symmetric structure (Figure 1(b)). Specifically, NC includes four key properties: (NC1) features from the same class collapse to their class mean; (NC2) class means are maximally separated and form a Simplex Equiangular Tight Frame (ETF); (NC3) feature means and classifier weights are mutually aligned as mirror images; and (NC4) classification reduces to a nearest-center decision rule. This structure is theoretically optimal for linear separability, minimizing intra-class variance and maximizing inter-class margin, and empirically correlates with low test error
under balanced settings.

However, under long-tail distributions, the NC structure is disrupted, leading to a common failure mode termed \textit{Minority Collapse}, where classifier weights for tail classes degenerate to nearly identical directions, inducing severe misclassification. As highlighted by \cite{inducingNC}, this is due to the imbalance in gradient contributions: majority classes dominate both attraction (intra-class compactness) and repulsion (inter-class separation) terms, while minority classes contribute negligibly. Consequently, classifier weights of tail classes are overwhelmed by repulsion from head classes, and their updates deviate from the NC geometry. The resulting biased decision boundaries favor majority classes and suppress minority accuracy.

To mitigate this, recent studies have attempted to reconstruct the ETF structure (NC1 and NC2), either by manually producing classifier weights \cite{inducingNC}, class prototypes \cite{BCL}, handcrafted ETF structure \cite{distribution_alignment}, or reshaping the feature space into a more stable configuration \cite{rotated_classidier, ARB, AISTATS}. However, these approaches overlook an important question: \textit{To what extent existing methods achieve feature-classifier alignment (NC3), and whether explicitly promoting this alignment is beneficial?} Intuitively, enforcing this alignment is helpful for further inducing NC in long-tail learning, which will encourage a better representation learning and provide a extra performance improvement.

To answer this, we study the space misalignment between feature and classifier vectors in long-tailed settings. As visualized in Figure~\ref{fig:toy_examples}(c), models trained with cross-entropy loss fail to achieve alignment, in contrast to the balanced case in Figure~\ref{fig:toy_examples}(b). We further observe that the similarity between these spaces correlates strongly with performance: 
Compared to baseline (purple curve), applying a constraint that further reduces this alignment (red curve) leads to consistent performance degradation. This suggests that space misalignment is not merely a by-product, but a core obstacle to robust representation learning under imbalance.

To formally analyze this, we introduce a geometric framework based on the Optimal Error Exponent (OEE) which is a classical information-theoretic measure that quantifies how quickly misclassification probability decays as the noise level decreases. We show that angular misalignment between feature and classifier directions provably slows convergence and deteriorates generalization. This analysis provides theoretical insight suggesting that even moderate misalignment can significantly deteriorate the generalization performance of the model.
Motivated by this, we propose three plug-and-play strategies to reduce space misalignment in standard long-tailed learning setups. As shown in Figure~\ref{fig:toy_examples}(d), our regularization leads to high alignment, and consistently improves both space similarity and classification accuracy (blue line in Figure~\ref{fig:toy_examples}(a)). Through extensive experiments on benchmarks, our approach achieves state-of-the-art performance while recovering stronger NC properties in both feature and decision spaces.

\section{Preliminaries}

The training set consists of $C$ classes, where the given dataset is balanced; each class contains $n$ samples and can be denoted as $\{(x_{i,c}, y_{i,c})\}$. Here, $x_{i,c}\in\mathbb{R}^d$ denotes the $i_{th}$ input sample of class $c$ and $y_{i,c}=c$ denotes its real label. The model is composed of a deep neural network. We can consider the layers before the classifier, which act as a feature extractor as a mapping ${h}$: $\mathbb{R}^d\to\mathbb{R}^p$ that outputs a $p$-dimensional feature vector ${h}(x)$. Following this, a linear classifier with weight matrix $\mathbf{W} \in \mathbb{R}^{C\times p}$ and biases $b\in \mathbb{R}^C$ takes the last-layer features as inputs and then outputs the class label. In detail, through classification scores via $f(x)=\mathbf{W}h(x)+b$, the predicted label is then given by $argmax_{c'}\langle  w_{c'},h\rangle +b_{c'}$, where $w_{c'}$ denotes the classifier weight for a specific class. Furthermore, we denote the class mean $\mu_c = \frac{1}{n}\sum_{i=1}^nh(x_{i,c})$ and the global mean $\mu_G = \frac{1}{C}\sum_{c=1}^C \mu_c$. 
\subsection{Simplex Equiangular Tight Frame (Simplex ETF)}
A general Simplex Equiangular Tight Frame (Simplex ETF) matrix $\mathbf{M}\in \mathbb{R}^{p\times C}$ is a collection of points in $\mathbb{R}^C$ specified by the columns of:
\begin{align}
    \mathbf{M} = \sqrt{\frac{C}{C - 1}}\, \mathbf{R} \left( \mathbf{I} - \frac{1}{C - 1} \mathbf{1}_C \mathbf{1}_C^\top \right),
\end{align}
where $\mathbf{I} \in \mathbb{R}^{C\times C}$
is the identity matrix and $\mathbf{1}_C \in \mathbb{R}^{C\times 1}$ is the ones vector, and $\mathbf{R}\in \mathbb{R}^{p\times C} (p\ge C)$ is a rotation orthogonal matrix ($\mathbf{R}^{\top} \mathbf{R}=\mathbf{I}$). $\mathbf{M}:= \{m_1, m_2,m_3...,m_C\}\in \mathbb{R}^{p\times C}$includes C classes with the $m_c$ weight.  

\subsection{Neural Collapse}
In the terminal phase of training on balanced datasets, as shown in Figure~\ref{fig:toy_examples}(b), 
it can be observed that the last-layer features will converge to class means, which in turn align with classifier weights, all forming the vertices of a symmetric simplex ETF \cite{Neural_Collapse}. Specifically, Neural Collapse (NC) can be formally described within 4 phases as follows:

\paragraph{(NC1) Within-class variability collapse} As training progresses, the features belonging to the same class collapse to their class means. Mathematically, this means the within-class covariance matrix $\Sigma_W$ approaches zero, that is:
\begin{align}
&\Sigma_W    = \frac{1}{Cn} \sum_{c=1}^{C} \sum_{i=1}^{n} \left( h(x_{i,c}) - \mu_c \right) \left( h(x_{i,c}) - \mu_c \right)^\top, \notag \\
  &\Sigma_W\to 0,
\end{align}
\paragraph{(NC2) Convergence to a simplex ETF} The mean vectors of each class converge to a simplex ETF:
\begin{align}
&\|\mu_c-\mu_G\|_2 - \|\mu_{c'}-\mu_G\|_2 \to 0, \notag \quad \forall c,c'\\
&\langle\hat{\mu}_c,\hat{\mu}_{c'}\rangle \to \frac{C}{C-1}\delta_{c,c'} - \frac{1}{C-1}, \quad \forall c,c'
\end{align}
where $\hat{\mu_c}=(\mu_c-\mu_G)/\|\mu_c-\mu_G\|_2$ denotes the renormalized class means and $\delta_{c,c'}$ is the Kronecker delta symbol. 
\paragraph{(NC3) Convergence to self-duality} The classifier weight vectors $\mathbf{W}_c$ become aligned with the class mean vectors $\mu_c$, reinforcing the global simplex ETF structure in both feature and decision spaces:
\begin{align}
    \left\| \frac{\mathbf{W^\top}}{\|\mathbf{W}\|_F}-\frac{\mathbf{\dot{M}^\top}}{\|\mathbf{\dot{M}}\|_F}\right\|_F \to 0,
\end{align}
where $\dot{M}=[\mu_c-\mu_G, c=1,...,C] \in \mathbb{R}^{p\times C}$.  
\paragraph{(NC4) Simplification to nearest center} The neural network classifier converges to a nearest class center classifier:
\begin{align}
    \arg\max_{c'} \langle {w}_{c'}, 
    {h} \rangle + b_{c'} \;\to\; \arg\min_{c'} \|{h} - {\mu}_{c'}\|_2.
\end{align}

\section{Space Misalignment Under Long-tail Setting}

When studying Neural Collapse (NC), the original theory \cite{Neural_Collapse} identified NC1 and NC2 as prerequisites for achieving alignment between the feature space and classifier vector space (NC3). However, in long-tail settings, the imbalanced sample sizes across classes induce skewed gradient signals, causing the phenomenon known as Minority Collapse \cite{Miniority_collapse}, which damages the ETF geometric structure. 
To address this challenge, existing methods have made notable progress in restoring the simplex structure. Nevertheless, as illustrated in Figure~\ref{experiment_results}, we still observe space misalignment in these studies, i.e., low cosine similarity between feature means and classifier weights. This motivates us to investigate the detrimental effects of space misalignment, even if the decision and feature spaces approximately approach the ideal ETF structure.

\subsection{Optimal Error Exponent Under Perfect Alignment}

\subsubsection{Setting}
To focus on and better quantify the harm that space misalignment might cause, following the settings of \cite{Neural_Collapse}, we construct our problem under an idealized condition where feature and classifier vector spaces already both converge to the simplex ETF. Assume we are given an observation: $h=\mu_c+z\in \mathbb{R}^C;z\sim \mathcal{N}(0, \sigma^2\mathbf{I})$ and $c \sim\{1,...,C\}$ denotes the unknown class index, independently distributed from z. We use a linear classifier, $\mathbf{W}h(x)+\mathbf{b}$ where weights $\mathbf{W}=[w_c:c=1,...,c]\in \mathbb{R}^{C\times C}$ and biases $\mathbf{b}=b_c\in \mathbb{R}^C$. Our decision rule is $\hat{\gamma}({h}) = \hat{\gamma}({h}; \mathbf{W}, \mathbf{b}) = \arg\max_c \langle w_c, {h} \rangle + b_c$.

\subsubsection{Optimal Error Exponent (OEE)}

Following \cite{Neural_Collapse}, to quantify classification performance theoretically, we consider the large-deviations error exponent. This exponent shows how quickly the misclassification probability decays as conditions become ideal (e.g., as noise or classification difficulty is reduced).
It is worth noting that in classification tasks (including the case of long-tail distributions), the training process drives the training loss to almost zero. Moreover, according to the research in \cite{NC1}, NC1 (within-class variance collapse) occurs in both of these situations. Therefore, here, leveraging the large-deviation theory is appropriate to analyze misclassification error and explain poor generalization performance.

\newtheorem{theorem}{Theorem}
\begin{theorem}[Large-Deviations Error Exponent]
If modeling classification with a bit of noise, the error exponent is defined as:
\begin{align}
    \beta(\mathbf{M}, \mathbf{W}, b) = \lim_{\sigma \to 0} -\sigma^2 \log P_{\sigma} \left\{ \hat{\gamma}(\mathbf{h}) \neq \gamma \right\}
\end{align}
where $\sigma$ represents the noise level. 
\end{theorem}

For very small noise, the misclassification probability $P_{\sigma}$ typically behaves like $\exp(-\beta/\sigma^2)$; \textit{a larger $\beta$ means the error probability decreases \textbf{exponentially faster} as the task gets easier} (noise $\sigma \to 0$). The error exponent thus captures the geometric separability of classes: if classes are well separated in feature space, the error of an ideal classifier will decrease rapidly (high $\beta$), while if some classes are too close (misaligned), the error decreases slowly (low $\beta$).
Furthermore, as shown in Theorem 2 proved by \cite{Neural_Collapse}, the maximum possible error exponent (over all possible arrangements of class feature means in a given dimension) is achieved when the class means are arranged as a centered simplex ETF (which essentially represents the most symmetric configuration).

\newtheorem{theorem2}{Theorem}
\begin{theorem}[Optimal Error Exponent (OEE)]
When both class means and classifier weights form a renormalized Simplex ETF, with the classifier bias being $0$:
\begin{align}
   \beta^* &= \max_{\mathbf{M}, \mathbf{W}, \mathbf{b}} \beta(\mathbf{M}, \mathbf{W}, \mathbf{b}) \quad \text{s.t.} \quad \|\mu_c\|_2 \leq 1 \quad \forall c \notag,\\
   \beta^* &= \beta(\mathbf{M^*}, \mathbf{M^*}, 0)\notag,\\
   \beta^* &= \frac{C}{C - 1} \cdot \frac{1}{4},
\end{align}
where $\mathbf{M^*}$ is a $C\times\ C$ matrix constructed by renormalized class mean $\hat{\mu_c}$ where $\hat{\mu_c}=(\mu_c-\mu_G)/\|\mu_c-\mu_G\|_2$. 
\end{theorem}

\subsection{Impact of Misalignment}

\subsubsection{Setting}

\begin{figure}[H]
\centering
\parbox{0.75\linewidth}{
    \centering
    \includegraphics[width=\linewidth]{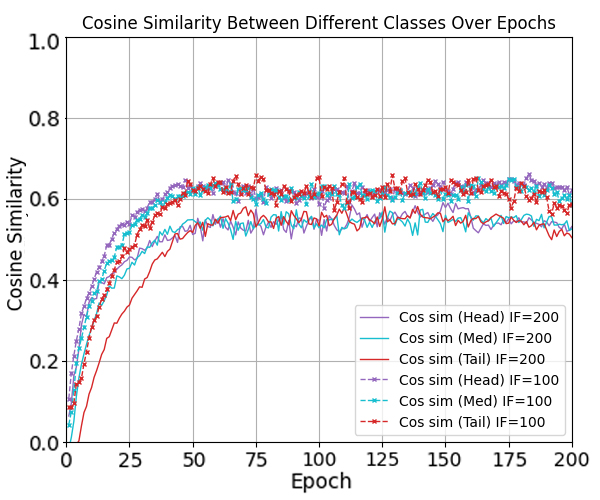}\\
    
}
\caption{Cosine similarity between the feature space and the decision space across $d$ classes under two imbalance factors (IF=$100$ and IF=$200$). 
}
\label{cossim}
\end{figure}

To quantify the influence of space misalignment, we follow the conventional setting in the original NC theory \cite{Neural_Collapse}, where both the feature and weight vector space converge to the Simplex ETF, i.e., $\|\hat{\mu_c}\|_2=\|\hat{w_c}\|_2=1; \langle \hat{\mu_c},\hat{\mu_{c'}}\rangle=\langle \hat{w_c},\hat{w_{c'}}\rangle=-\frac{1}{C-1}, (c\neq c')$.
Moreover, we consider a simplified but insightful case: \textit{uniform angular misalignment}. This setting assumes that each classifier weight vector $w_c$ forms a fixed angle $\alpha \in [0,\frac{\pi}{2}]$ with its corresponding feature mean $\mu_c$ such that 
$$
\langle \hat{\mu_c},\hat{w_c}\rangle = \cos \alpha, \forall c.
$$
That is, the classifier weights can be obtained by a single isoclinic rotation R 
(\(
R = \cos\alpha\,I + \sin\alpha\,A
\)
with \(A^\top = -A\ ,A^2=-I\). 
We discuss an even feature dimension here to guarantee the existence of a real orthogonal $A$ here; Odd-dimension features can always be zero-padded to satisfy this condition without altering the analysis results.) 
from the unit-norm feature simplex.
Actually, this uniform angular misalignment can also be observed from an empirical observation as shown in Figure~\ref{cossim}, where the misalignment angles across head, medium, and tail classes are approximately consistent as training progresses.

Denoting $v=\hat{w_c}-\hat{w_{c'}}; d=\hat{\mu_c}-\hat{\mu_{c'}}$, we have the following Theorem 3 for a unit-norm standard simplex. Theorem 3 provides the distance and inner product expressions corresponding to the symmetric structure among the class centers, offering key constants for the explicit expression of the error exponent in subsequent steps. It can be directly applied to simplify the error comparison in both spatially aligned and unaligned scenarios.

\begin{theorem}[Standard Simplex ETF Distance Properties]
Let $\{\mu_c\}_{c=1}^C \subset \mathbb{R}^p$ from a  standard Simplex ETF, we have:
\begin{align}
    d^\top\mu_c=\frac{1}{2}\|d\|^2,\quad \|d\|^2=\frac{2C}{C-1}
\end{align}
\end{theorem}

\subsubsection{Proof Sketch}
In this proof, we aim to quantify how a uniform misalignment angle $\alpha$ affects the Optimal Error Exponent (OEE). The proof proceeds in the following key steps: 
\textbf{Step 1: Problem Reformulation}\quad We first express the misclassification probability as the chance that a Gaussian perturbation $z$ causes the inequality $\langle \hat{w_c},h\rangle  \leq  \langle \hat{w_{c'}},h\rangle$. Then, applying Lemmas 1 and 2, we convert this probability into a constrained optimization problem and get the closed form of this problem. 
\textbf{Step 2: Pairwise Error Exponent Derivation} \quad We then introduce the misalignment angle $\alpha$ and use the geometric properties of ETF (Lemma 3) to derive the pairwise error exponent under the simplex alignment setting (Theorem 4).
\textbf{Step 3: OEE Upper Bound} \quad Finally, we derive the upper bound for the OEE under the simplex misalignment setting (Theorem 5), demonstrating that even with perfect ETF geometry, a common rotation degrades the OEE quadratically in $\cos\alpha$. \textbf{All the detailed proof contents in this part can be found in the Appendix.}

\subsubsection{Problem Reformulation}

Considering the fundamental tool from the Large-Deviations Theory \cite{large_deviation_theory} in Lemma 1, we transform the task of calculating the misclassification into the Minimum-Norm optimization problem.
\newtheorem{lemma}{Lemma}
\begin{lemma}
    Suppose $\mathcal{K}$ is a closed set and $0\notin \mathcal{K}$. Then as $\sigma \to 0$:
    \begin{align}
        -\sigma^2 \log P_{\sigma} \{ z \in \mathcal{K} \} \to \min \left\{ \frac{1}{2} \|z\|_2^2 : z \in \mathcal{K} \right\}
    \end{align}
\end{lemma}

Based on Lemma 1, the probability of misclassification of class $c$ to $c'$ can be transformed into an optimization problem, and then obtain in a closed form.
\begin{lemma}
    For any two distinct classes $c\neq c'$, following the previous setting $v=\hat{w_c}-\hat{w_{c'}}; d=\hat{\mu_c}-\hat{\mu_{c'}}$, the large-deviations error exponent for misclassifying $c\to c'\quad \beta_{c,c'}=-\lim_{\sigma \to 0}\sigma^2 \log P \left\{ \hat{\gamma}(\mathbf{h})=c'| \gamma=c\right\}$ admits the closed form:
    \begin{align}
        \beta_{c,c'}=\frac{(v^\top\hat{\mu_c})^2}{2\|v\|^2}
    \end{align}
\end{lemma}

\subsubsection{Pairwise Error Exponent Derivation}

To analyze how space misalignment affects the error exponent, the misalignment angle setting is introduced in Lemma 3.

\begin{lemma}
\label{lem:cross}
Let $\{\mu_j\}_{j=1}^{C}\subset\mathbb R^{D}$ be unit vectors forming a
regular simplex, i.e.\
\[
\|\mu_j\| = 1, \qquad
\mu_j^{\!\top}\mu_{j'}=
  \begin{cases}
    1, & j=j',\\[2pt]
   -\dfrac{1}{C-1}, & j\neq j'.
  \end{cases}
\]
Apply a common orthogonal transform
\(
R = \cos\alpha\,I + \sin\alpha\,A
\)
with \(A^\top = -A\ ,A^2=-I\) to obtain the “mis‑aligned” vectors
\(w_j = R\mu_j\) and assume the
\textit{unified misalignment angle}
\(\mu_j^{\!\top}w_j = \cos\alpha\) for every \(j\).
Then for any fixed class \(c\) and all \(c'\neq c\):
\begin{align}
\hat{\mu_c}^{\!\top}\hat{w_{c'}}
 = -\frac{\cos\alpha}{C-1} + \zeta_{c,c'},
 \quad
 \zeta_{c,c'} = \sin\alpha\;\hat{\mu_c}^{\!\top} A \hat{\mu_{c'}} .
\end{align}
The residuals \(\zeta_{c,c'}\) satisfy
\begin{align}
\quad
\sum_{c'\neq c} \zeta_{c,c'} = 0 .
\end{align}
\end{lemma}

Based on the result of Lemma 3 that provides the inner product expression between $\hat{\mu_c}$ and $\hat{w_{c'}}$, we can substitute these results into the closed form expression in Lemma 2 to derive the pairwise error exponent with misalignment angle $\alpha$.
\begin{theorem}[Error Exponent under simplex misalignment setting]
\label{thm:beta}
For any two distinct classes \(c\neq c'\), 
\(v = \hat{w_c} - \hat{w_{c'}}\), \( d=\hat\mu_c-\hat\mu_{c'}\).  Then
\(\|d\|^{2} =\|v\|^2= 2C/(C-1)\)
and the Chernoff‑type error exponent is
\begin{align}
\beta_{c,c'} \;=\;
 \frac{C-1}{4C}\,
 \Bigl(
   \cos\alpha\bigl(1+\tfrac1{C-1}\bigr)
   - \zeta_{c,c'}
 \Bigr)^{2}
\end{align}
where $\zeta_{c,c'} = \sin\alpha\,\hat{\mu_c}^{\!\top}A\hat{\mu_{c'}}$. When the space misalignment angle is $0$ (aligned), the same form of Optimal Error Exponent under perfect alignment in Theorem 2 can be obtained \cite{Neural_Collapse}.
\end{theorem}

\subsubsection{OEE Upper Bound}

We aim to quantify how such misalignment influences the large-deviations error exponent $\beta$. Under this setup, the decision boundary between any two classes $c$ and $c'$ depends on the angle between their projected feature means, i.e., $\langle \hat{\mu_{c'}},\hat{w_c}\rangle$. Because the classifier is misaligned, the effective projection of the feature mean onto its weight vector is shrunk by a factor of $\cos \alpha$.
Therefore, the margin between any two classes scales by $\cos \alpha$.
As such, we connect the OEE between the perfect alignment and the space misalignment case by using Theorem 5, where the detailed proof can be found in the Supplementary.

\begin{theorem}
[Optimal Error Exponent Under Simplex Misalignment Setting] 
Following \cite{Neural_Collapse}, the optimal error exponent quantifies the asymptotic rate at which the misclassification probability decays as the noise level becomes relatively small. Given a space misalignment angle $\alpha$, optimal error exponent is formulated as:
\begin{align}
    \beta^{*'} =\min_{c'\neq c}\beta_{c,c'} \le \frac{1}{4}\cos^2\alpha \frac{C}{C-1}=\cos^2\alpha \beta^*.
\end{align}
\end{theorem}

This result illustrates that even if the geometry of the feature space is perfectly optimal (e.g., a simplex ETF), angular misalignment between features and classifiers can substantially impair the theoretical error rate decay. As such, reducing the angle of misalignment is crucial to induce NC under long-tail learning.

\section{Proposed Methods}

\begin{figure*}[!t]
  \centering
  \includegraphics[width=0.7\linewidth]{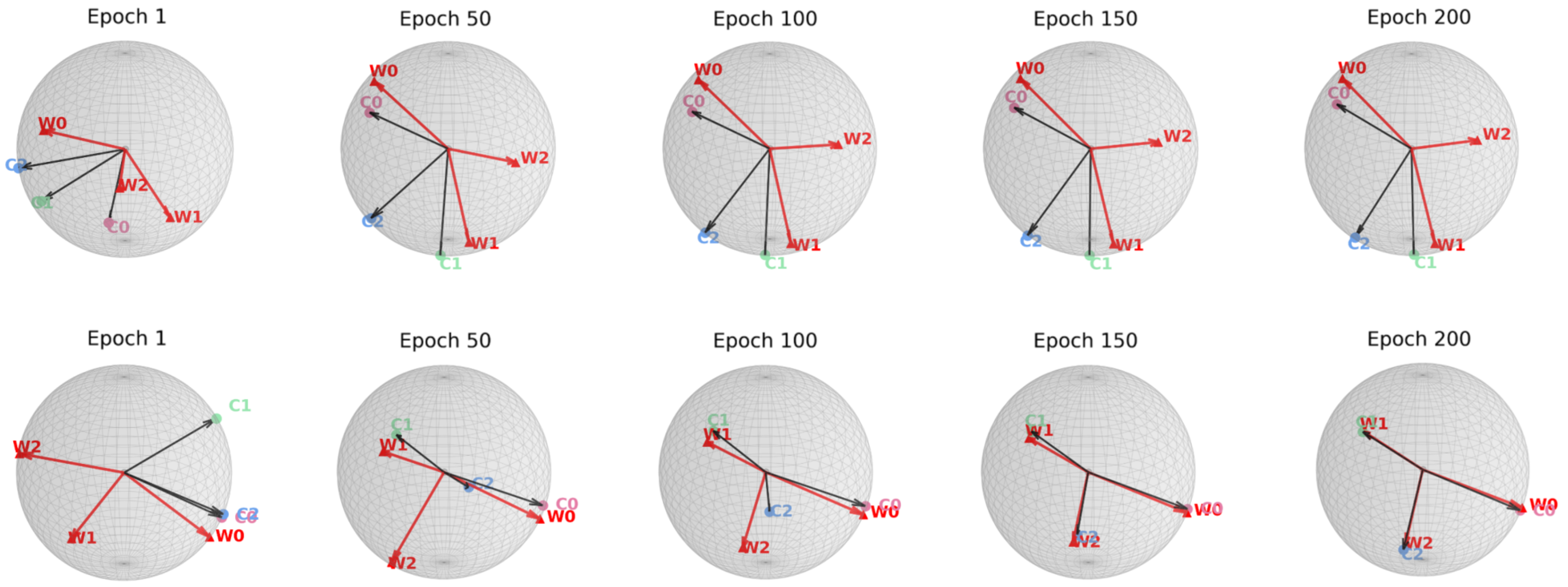}\hfill
  \caption{A toy example illustrating the process of space alignment using our proposed SpA-Reg method. Each sphere shows the change of orientations of classifier weights (red arrows) and class feature means (black arrows) as training progressed. The top row visualizes the standard long-tail learning, i.e., training with cross-entropy loss, where significant misalignment between the classifier weights and the feature center persists throughout the whole training process. In contrast, for the bottom row, the classifier weights gradually aligned with the feature mean during training.
  }
  \label{fig:ten-grid}
\end{figure*}

To verify the importance of space alignment, we propose three methods for explicitly aligning the feature space and the decision space, which enable a convenient integration with current methods.

\paragraph{Similarity regularization (SpA-Reg)} 

A straightforward way is to improve the cosine similarity between the feature and the classifier vector space by applying a regularization term:
\begin{equation}
\mathcal{L}_{SpA-Reg} = \frac{1}{C}\sum_{c=1}^C [1-cos(\hat{w_c},\hat{\mu_c})],
\end{equation}
which can be added to the main loss with a scaling hyperparameter $\lambda$:
\begin{equation}
\mathcal{L}_{total}=\mathcal{L}_{main} + \lambda\mathcal\cdot \mathcal{L}_{SpA-Reg}.
\end{equation}
A toy example is provided in Figure~\ref{fig:ten-grid} to show the effectiveness of directly correcting the misalignment in standard long-tail learning. 

\paragraph{Spherical linear interpolation (SpA-SLERP)}

Inspired by the spherical geometry of NC, this approach rotates the classifier weights towards their corresponding class feature mean vectors, which naturally reduces the misalignment angle.
During training, given a predefined threshold $\tau$, when the cosine similarity, $\cos sim=\langle \hat{w_c},\hat{\mu_c} \rangle$, is below this threshold, an SLERP algorithm will execute in the following form:
\[
w_c \leftarrow \|w_c\| \left( 
\frac{\sin((1 - \alpha_t)\theta_c)}{\sin \theta_c} \hat{w}_c +
\frac{\sin(\alpha_t \theta_c)}{\sin \theta_c} \hat{\mu}_c 
\right).
\]
Here, we apply a cosine schedule to the interpolation coefficient $\alpha$ to realize a more flexible optimization and reduce its interference in the early stage of training:
\[
\alpha_t=\alpha_{max}\frac{1-\cos (\frac{\pi t}{T})}{2},
\]
where $t$ denotes the current training epoch and $T$ denotes the total number of epochs.

\paragraph{Gradient projection (SpA-Proj)}

To prevent the classifier weights from deviating excessively from the mean values of the corresponding class features, we propose a gradient projection mechanism that selectively adjusts the update direction of classifier weights. 
After calculating the gradient $g_c=\nabla_{w_c}\mathcal{L} $, we decompose the gradient into a radial and a tangential part:
\begin{align}
    g_{rad}=\langle g_c,\hat{w_c}\rangle  \hat{w_c} \quad g_{tan}=g_c-g_{rad}.
\end{align}
Then, we project the renormalized class mean $\hat{\mu_c}$ onto the same tangent space:
\begin{align}
    d_c=\hat{\mu_c}-\langle g_c,\hat{w_c}\rangle  \hat{w_c}, \quad \hat{d_c}=\frac{d_c}{\|d_c\|}.
\end{align}
Moving along $d_c$ is the steepest way to decrease the misalignment angle. Based on this, we can remove the harmful gradient component during training. Denote $p_c=\langle g_{tan}, \hat{d_c\rangle}$, we suppress the gradient component that pushes the classifier weights away from the class means:
\begin{align}
    g^{\mathrm{safe}}_{\mathrm{tan}}=
\begin{cases}
g_{\mathrm{tan}}-p_c\hat d_c, & p_c>0,\\[2pt]
g_{\mathrm{tan}}, & p_c\le 0.
\end{cases}
\end{align}
Then we mix the safe direction with the original gradient to maintain the general training signal:
\begin{align}
    g_{final} = (1-\gamma) g^{\mathrm{safe}}_{\mathrm{tan}} +\gamma g_c,
\end{align}
where $\gamma$ is a soft projection coefficient.

\section{Experiments}

\paragraph{Datasets} 
Following the standard long-tail evaluation process \cite{GLMC, ARB, distribution_alignment, inducingNC}, we use the modified long-tail version of CIFAR-10 \cite{CIFAR10}, CIFAR-100 \cite{CIFAR10}, and ImageNet \cite{imagenet} dataset. In specific, we leverage the imbalance ratio $q$ (defined by the ratio of the samples between the most-frequent class and the rarest class in the whole dataset: $q=N_{max}/N_{min}$) and the exponential decay to create a long-tail dataset with different degrees of imbalance. For CIFAR10-LT and CIFAR-100-LT, we modify each of them with three imbalance factors \{200,100,50\}. For ImageNet-LT, it has an imbalance ratio of 256 with 1000 different classes. We train each model on the imbalanced training set and evaluate it in the balanced validation/test set.

\paragraph{Implementation}
All experiments are conducted on NVIDIA GeForce RTX 4090 GPUs by using PyTorch. Following \cite{GLMC,distribution_alignment,ARB}, ResNet-32 \cite{ResNet32} is applied on both CIFAR10-LT and CIFAR100-LT, ResNet-50 \cite{ResNet32} and ResNet-50-32x4d \cite{ResNext} on ImageNet-LT. All models are optimized by applying the SGD optimizer with a momentum of 0.9. 
When our methods are incorporated into the baseline methods, all hyperparameters and learning rate scheduling strategies will follow those of the respective baselines. All experiments are repeated with three different random seeds; the reported results are the average over these runs. \textbf{More details of hyperparameters in our methods are shown in the Appendix.}

\paragraph{Baselines}
 We choose baseline methods with fundamentally different motivations. Except for standard training with cross-entropy (CE) loss, the chosen strategies include contrastive learning and data augmentation-based strategies: KCL \cite{KCL}, TSC \cite{TSC}, HCL \cite{HCL}, GLMC \cite{GLMC}. Meanwhile, we also consider two-stage methods: BBN \cite{BBN}, RIDE \cite{RIDE}, MaxNorm \cite{MaxNorm}. Moreover, recent approaches motivated by the Neural Collapse (NC) phenomenon, e.g., INC \cite{AISTATS}, fixed classifier as ETF \cite{inducingNC}, RBL \cite{rotated_classidier}, ARB \cite{ARB}, DisA \cite{distribution_alignment}, are also taken into consideration.

\begin{table}[t]
\centering
\resizebox{\linewidth}{!}{

\begin{tabular}{lcccccc}
\toprule
\multirow{2}{*}{\textbf{Method}} & \multicolumn{3}{c}{\textbf{CIFAR-10-LT}} & \multicolumn{3}{c}{\textbf{CIFAR-100-LT}} \\
\cmidrule(lr){2-4} \cmidrule(lr){5-7}
 & \textbf{200} & \textbf{100} & \textbf{50} & \textbf{200} & \textbf{100} & \textbf{50} \\
\midrule
BBN & / & 79.9 & 82.2 & / & 42.6 & 47.1 \\
KCL & /& 77.6 & 81.7 & / & 42.8 & 46.3  \\
TSC & /& 79.7 & 82.9 & /  & 43.8 & 47.4 \\
HCL & /& 81.4 & 85.4 & / & 46.7 & 51.9  \\
MiSLAS & /& 82.1 & 85.7 & / & 47.0 & 52.3  \\
RIDE (3 experts) & / & 81.6 & 84.0 & /& 48.6 & 51.4  \\
RBL & 81.2 & 84.7 & 87.6 & 48.9 & 53.1 & 57.2 \\
INC-DRW & 75.8 & 81.9 & 82.7 & 42.5 & 48.6 & 51.7 \\
\midrule
CE* & 70.1 & 75.4 & 78.3 & 38.5 & 42.1 & 48.1 \\
CE*+\textbf{SpAReg} & 71.9 & 76.3 & 79.3 & 39.9 & 44.3 & 49.2 \\
CE*+\textbf{SpASLERP} & 71.6 & 76.5 & 79.0 & 39.1 & 43.1 & 48.9 \\
CE*+\textbf{SpAProj} & 71.7 & 76.6 & 79.1 & 39.4 & 43.3 & 48.6 \\
\midrule
ETF-DR & 71.9 & 76.5 & 81.0 & 40.9 & 45.3 & 50.4 \\
ETF-DR+\textbf{DisA} & 73.7 & 78.5 & 81.4 & 41.5 & 45.9 & 51.1 \\
ETF-DR+\textbf{SpAReg} & 73.0 & 79.0 & 82.3 & 41.6 & 46.3 & 50.9 \\
\midrule
ARB & 79.6 & 83.3 & 85.7 & 44.5* & 47.2 & 52.6 \\
ARB+\textbf{SpAReg} & 81.2 & 84.0 & 86.7 & 45.6 & 51.1 & 55.2 \\
ARB+\textbf{SpASLERP} & 80.7 & 83.8 & 86.6 & 44.7 & 49.8 & 54.2 \\
ARB+\textbf{SpAProj} & 81.0 & 84.2 & 86.4 & 45.1 & 49.8 & 54.1 \\
\midrule
GLMC & 83.4*& 87.8 & 90.2  & 50.8* & 55.9 & 61.1 \\
GLMC+MaxNorm (two-stage) & /  & 87.6& 90.2 & / & 57.1& 62.3 \\
GLMC+\textbf{SpAReg} & 83.9 & 88.5& \textbf{91.1}  & 52.0 & 58.2 & 63.5 \\
GLMC+\textbf{SpASLERP} & 83.9 & 88.8  & 90.9& 52.1 & 58.0 & 63.3 \\
GLMC+\textbf{SpAProj} & \textbf{84.2} & \textbf{88.9} & 90.7 & \textbf{52.2} & \textbf{58.3} & \textbf{63.6} \\
\bottomrule
\end{tabular}
}
\caption{Long-tailed classification accuracy (\%) with {ResNet-32} under imbalance ratios \{200,100,50\} on CIFAR-10-LT and CIFAR-100-LT. The methods or results marked with (*) denote the reproduced result by ourselves.}
\label{tab:cifar10_100_results}
\end{table}

\subsection{Long-tailed Benchmark Results}

\textbf{All results with standard deviation} and \textbf{accuracies on three splits of the set of classes}: Many-shot (more than 100 images), Medium-shot (20-100 images)
and Few-shot (less than 20 images), \textbf{are shown in the Appendix}.

\paragraph{CIFAR10-LT and CIFAR100-LT}

Extensive experiments are conducted to demonstrate the effectiveness of aligning the feature and classifier vector space during training. Table~\ref{tab:cifar10_100_results} reports the accuracy of various methods on CIFAR-10-LT and CIFAR-100-LT with three imbalance ratios: 50, 100, and 200. Our approach achieves an improvement in accuracy ranging from 0.5 to 2.6.
Moreover, we can observe that, compared to CIFAR-10-LT, when applying our proposed methods on CIFAR-100-LT, the baseline methods can have a higher performance gain than on CIFAR-10-LT. The main reason is that under a long-tail setting, the increase in the number of classes exacerbates the space misalignment, thereby making the effect of space alignment more significant.  

\begin{figure}[!t]
\centering
\parbox{0.495\linewidth}{
    \centering
    \includegraphics[width=\linewidth]{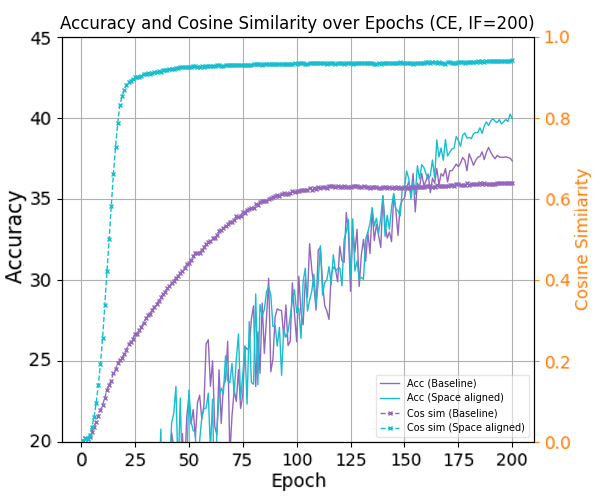}\\
    (a) 
}
\parbox{0.495\linewidth}{
    \centering
    \includegraphics[width=\linewidth]{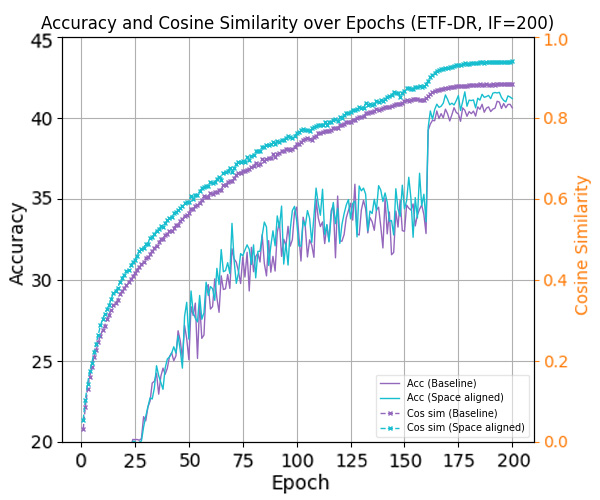}\\
    (b) 
}
\hfill
\parbox{0.495\linewidth}{
    \centering
    
    \includegraphics[width=\linewidth]{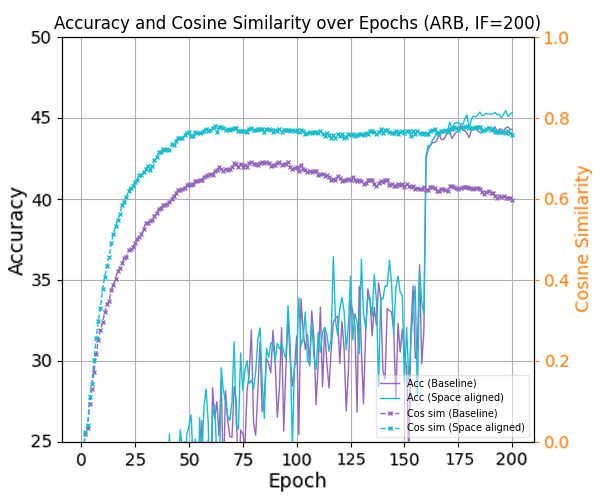}\\
    (c)  
}
\hfill
\parbox{0.495\linewidth}{
    \centering
    \includegraphics[width=\linewidth]{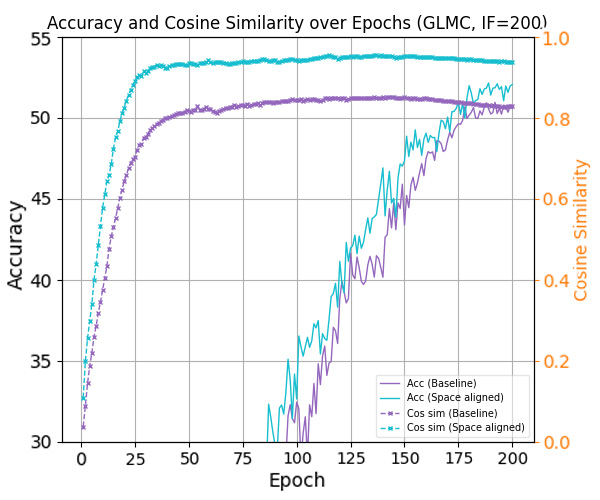}\\
    (d)  
}
\caption{Performances and corresponding
cosine similarities between class feature means and classifier
weights on the CIFAR-100 dataset with the imbalance factor of 200. 
}
\label{experiment_results}
\end{figure}

As shown in Figure \ref{experiment_results}, our space alignment strategies lead to high alignment during training, improving the classification performance consistently at the same time. Meanwhile, we can observe that in the later stages of training, as the loss approaches zero and the model enters the small-noise regime, the model with a higher space alignment angle can have a better generalization performance. This observation matches our theoretical analysis: the optimal error exponent $\beta$ will reduce with respect to $\cos^2 \alpha$ when the model enters the large deviation regime. The experiment results confirm the validity of explicitly aligning the decision space and the feature space during training. Combining space alignment with different types of long-tail based methods can yield consistent performance improvement compared with original baselines and achieve the best performance.

\paragraph{ImageNet-LT} We further conduct more experiments with different types of long-tailed classification methods on the ImageNet-LT dataset. To ensure a fair comparison in the experiment, we use the ResNet-50 following \cite{ARB,inducingNC} and ResNeXt-50 following \cite{GLMC}. As the experiment results shown in Table \ref{imagenet}, through aligning the feature space and the decision space, our methods outperform related long-tail approaches and achieve the highest performance.

\begin{table}[!t]
\centering
\scriptsize
\begin{tabular}{lccc}
\toprule
\textbf{Method} & \textbf{Backbone} & \textbf{All} \\
\hline
KCL                &    ResNet-50   & 51.5 \\
TSC                &    ResNet-50     & 52.4 \\
MiSLAS             &      ResNet-50    & 52.7 \\

LDAM-DRW           &     ResNet-50    & 47.7 \\
LDAM-DRW+DisA      &    ResNet-50     & 48.5 \\
CE-DRW             &    ResNet-50     & 47.1 \\
\hline
CE*                        & ResNet-50    & 44.3 \\
CE*+\textbf{SpAReg}                 & ResNet-50    & 44.8 \\
CE*+\textbf{SpASLERP}               & ResNet-50    & 44.7 \\
CE*+\textbf{SpAProj}                & ResNet-50    & 44.8 \\
ETF-DR                     & ResNet-50    & 44.7 \\
ETF-DR+\textbf{SpAReg}              & ResNet-50    &  45.3\\ 
ARB                        & ResNet-50    &   52.8 \\ 
ARB+\textbf{SpAReg}                 & ResNet-50    & 53.2 \\ 
ARB+\textbf{SpASLERP}               & ResNet-50    &  53.1\\ 
ARB+\textbf{SpAProj}                & ResNet-50    & \textbf{53.6} \\ 
\midrule
CE-DRW                     & ResNeXt-50   & 46.4 \\
CE-LWS                     & ResNeXt-50   & 47.7 \\
LADE                       & ResNeXt-50   & 53.4 \\
RBL                        & ResNeXt-50   & 53.5 \\
INC-DRW                    & ResNeXt-50   & 53.0 \\
INC-DRW-cRT                & ResNeXt-50   & 54.5 \\
\hline
GLMC & ResNeXt-50  & 56.3 \\
GLMC+\textbf{SpAReg}& ResNeXt-50  & \textbf{56.7} \\ 
GLMC+\textbf{SpASLERP}& ResNeXt-50   & 56.5 \\
GLMC+\textbf{SpAProj}& ResNeXt-50  & \textbf{56.7} \\
\bottomrule

\end{tabular}
\caption{Long-tailed classification accuracy (\%) on ImageNet-LT. The method or result marked with (*) denotes the reproduced result by ourselves.}
\label{imagenet}
\end{table}

\section{Related Work}
This class imbalance leads to deep models favoring the head classes during training and having difficulty achieving good generalization performance on the tail classes. Classic solutions include data resampling (oversampling the tail classes or undersampling the head classes) \cite{chu2020feature,shen2016relay}, and cost-sensitive re-weighting based on class weights to enhance the loss gradient weights of the tail classes \cite{cao2019learning, khan2017cost}. However, aggressive resampling methods may disrupt the original data distribution: oversampling may lead to overfitting of the tail classes, while undersampling or high loss weights may compromise the overall accuracy \cite{GLMC}. 

Recent works inspired by Neural Collapse (NC) have explored how to improve long-tailed learning by promoting NC geometry. One line of methods explicitly induces NC patterns via regularization. For example, \cite{AISTATS} introduced a feature alignment term to cross-entropy loss to encourage intra-class feature collapse and inter-class orthogonality, effectively recovering NC1 and NC2 even under class imbalance. \cite{ARB} analyzed minority collapse from the gradient perspective and proposed ARB-Loss, which balances attraction and repulsion forces to stabilize classifier gradients and restore the NC structure. From a representation learning angle, \cite{BCL} proposed Balanced Contrastive Learning (BCL), which uses class-balanced sampling and averaging to prevent head-class dominance in contrastive learning, pushing the feature space toward a simplex ETF.

Another line of work imposes NC-friendly geometry through fixed or structured classifiers. \cite{inducingNC} argued that since the optimal classifier under NC is a simplex ETF, a learnable classifier may not be necessary. They fixed the classifier to a random ETF and only trained the feature extractor, leading to natural NC emergence even with imbalance. \cite{rotated_classidier} proposed Rotated Balanced Learning (RBL), adding a learnable rotation to align the feature space with the fixed classifier. \cite{distribution_alignment} proposed Distribution Alignment (DisA), using optimal transport to align features with a fixed ETF structure. This lightweight regularization improves compatibility with existing long-tail methods. \cite{yan2024neural} further extended NC by proposing Neural Collapse to Multiple Centers (NCMC), allowing each class to collapse to multiple prototypes, especially benefiting tail classes with enhanced feature diversity.

\section{Conclusion}
In this paper, we comprehensively discuss the phenomenon of space misalignment, an often overlooked problem when inducing Neural Collapse into long-tail learning. Through an analytical framework based on the Optimal Error Exponent, we quantified the detrimental effect of space alignment theoretically. Based on this theoretical insight, we proposed three plug-and-play alignment strategies that do not require architectural modifications to origin methods. Extensive experiments verify that these alignment strategies substantially enhance space similarity, achieving the state-of-the-art performances at the same time. Our findings emphasize the important role of space alignment when inducing NC to long-tail learning, offering novel perspectives for addressing data imbalance issues. 

\section*{Acknowledgement}
The work was partially supported by the following: 
WKU Internal (Faculty/Staff) Start-up Research Grant under No. ISRG2024009,
WKU 2025 International Collaborative Research Program under No. ICRPSP2025001.

\newpage
\bibliography{aaai2026}

\end{document}